%% file: main.tex
\pdfoutput=1

\documentclass[11pt]{article}

\input{packages}

\input{sections/00_titleauthor}

\begin{document}
\maketitle
\input{sections/00_abstract}
\input{sections/01_introduction}

\input{sections/02_methods}
\input{sections/03_experiments}
\input{sections/05_relatedworks}

\input{sections/04_discussion}

\bibliographystyle{acl_natbib}
\bibliography{anthology}

\clearpage
\appendix
\input{sections/Appendix}

\end{document}

%% file: packages.tex
\usepackage[]{EMNLP2023}

\usepackage{times}
\usepackage{latexsym}

\usepackage[T1]{fontenc}
\usepackage[utf8]{inputenc}
\usepackage{microtype}
\usepackage{lipsum}
\usepackage{multicol}
\usepackage{multirow}
\usepackage{inconsolata}
\usepackage{booktabs}       %
\usepackage{amsfonts}
\usepackage{graphicx}
\usepackage{amsmath}
\usepackage{tabularx}
\usepackage{subfigure}
\usepackage{booktabs}
\usepackage{enumitem}

\newcommand{\ra}[1]{\renewcommand{\arraystretch}{#1}}

\DeclareMathOperator*{\argmax}{argmax}
\DeclareMathOperator*{\argmin}{argmin}

%% file: sections/00_titleauthor.tex
\title{Future Lens: Anticipating Subsequent Tokens from a Single Hidden State}

\author{Koyena Pal \\
  Northeastern University \\
  \texttt{pal.k@northeastern.edu} \\\And
  Jiuding Sun \\
  Northeastern University \\
  \texttt{sun.jiu@northeastern.edu} \\
    \And
  Andrew Yuan \\
  UMass Amherst \\
  \texttt{awyuan@umass.edu} \\\AND
  Byron C. Wallace \\
  Northeastern University \\
  \texttt{b.wallace@northeastern.edu} \\\And
  David Bau \\
  Northeastern University \\
  \texttt{d.bau@northeastern.edu}}

%% file: sections/00_abstract.tex
\begin{abstract}

We conjecture that hidden state vectors corresponding to individual input tokens encode information sufficient to accurately predict several tokens ahead. 
More concretely, in this paper we ask: Given a hidden (internal) representation of a single token at position $t$ in an input, can we reliably anticipate the tokens that will appear at positions $\geq t+2$?
To test this, %
we measure linear approximation and causal intervention methods in GPT-J-6B to evaluate the degree to which individual hidden states in the network contain signal rich enough to predict future hidden states and, ultimately, token outputs.
We find that, at some layers, we can approximate a model's output with more than 48\% accuracy with respect to its prediction of subsequent tokens through a single hidden state. Finally we present a ``Future Lens'' visualization that uses these methods to create a new view of transformer states.
\end{abstract}

%% file: sections/01_introduction.tex
\section{Introduction}

Do hidden states in large language models (LLMs) encode tokens farther than a single token ahead? %
If so, how can we decode this sequence of tokens from %
a single state?
In this work we empirically investigate these questions using GPT-J-6B \cite{wang2021gpt}. %
We train models to predict hidden states several tokens ahead of a given position $t$ based \emph{only} on a contextualized representation of the input at this position. 

Auto-regressive transformer language models are typically trained to predict one token ahead, but %
recent work has hinted that individual hidden states may contain more information than just probabilities of the %
following %
token. 
For example, Meng \emph{et al.} \citeyearpar{meng2022locating} trace information flow from subject tokens to associated attribute predictions many steps ahead. 
Elsewhere, Gurnee \emph{et al.} \citeyearpar{gurnee2023finding} suggest that neurons in %
early layers %
are dense with information, %
while middle layers have dedicated neurons %
that represent high-level contextual features.

Other related %
efforts have %
passed hidden intermediate states directly to the decoder head (skipping in-between layers) to ``verbalize'' such embeddings   \cite{din2023jump, belrose2023eliciting, logit_lens}. 
Studies of memorization~\cite{carlini21extracting, carlini2023quantifying, thesecretsharer19} have identified the presence of very long memorized sequences generated by language models, and \citet{zhang2020accelerating} shows that progressively dropping layers during computation can still achieve a similar prediction output of the model when compared against their fully computed model run.

In this work we ask: To what extent can we extract information about future (beyond subsequent) tokens from a single hidden token representation? 
To answer this, we conduct three experiments. 
First, extending the ideas of Tuned Lens~\cite{belrose2023eliciting,din2023jump} and the Logit lens~\cite{logit_lens}, we train linear models to approximate future model predictions several tokens in the future, in order to reveal the extent to which individual hidden states may directly encode subsequent tokens. 
Second, we perform a causal intervention study in which we transplant individual hidden states from one context to a completely different context and measure the extent to which future tokens that were predicted in the original context can be predicted in the foreign context. 
Finally,  we fit a ``soft prompt'' to explicitly learn an optimal prompt that permits reading out information about subsequent tokens from a hidden state. Our code and data is available at \url{https://future.baulab.info}

%% file: sections/02_methods.tex
\section{Methods}
\label{section:methods}

To unveil the information about ``future'' tokens implicitly encoded in a single transformer state vector, we develop and compare several methods for predicting future tokens from a single hidden state. Each of our methods has the same goal: Extract accurate predictions of a model's probability distribution several tokens ahead, based on the information in only one hidden state at a single layer at one token of the transformer. 

For our evaluations we use an autoregressive transformer \citep{vaswani2017attention} language model %
defined as a function $G : X \rightarrow Y$ over vocabulary $V$ of size $|V| = d_{v}$. $G$ takes in a sequence of tokens $x = [x_1, ...., x_T] \in X, x_i \in V$ and maps %
this to a probability distribution $y_T \in Y \subset [0,1]^{d_{v}}$, which (greedily) predicts the next-token $x_{T+1} = \argmax y_T$. To generate %
additional tokens, the top predicted token $x_{T+1}$ is added to the sequence of tokens $[x_1, ...., x_T, x_{T+1}]$ and the process is repeated until the next $N$ tokens are produced.

To calculate each predicted probability distribution from an input sequence $x$, the transformer %
performs a sequence of computations at $L$ layers; %
this can be decomposed as:
\begin{align}
    G(x) = D(b_L( \cdots (b_2(b_1(E(x)))) \cdots ))
\end{align}
Where the first step $E:\rightarrow \mathbb{R}^{d_{h}}$ embeds each input token into an initial hidden representation, $e(x_i) = h_{i}^{0} \in \mathbb{R}^{d_{h}}$; each layer $b_{l}: \mathbb{R}^{d_{h}\times T} \rightarrow \mathbb{R}^{d_{h}\times T}$ transforms the sequence of representations; and the decoder $D: \mathbb{R}^{d_{h}} \rightarrow Y$ decodes the predicted probability distribution $y_T = D(h_{T}^{L})$ from the last layer at the last token. We %
write the output of layer $l$ as $H_{l} = b_{l}(H^{l-1})$, where:
\begin{align}
    H^{l} = (h_1^l, ..., h_T^l) \in \mathbb{R}^{d_h \times T}
\end{align}

\begin{figure}
    \centering
\includegraphics[width=\columnwidth]{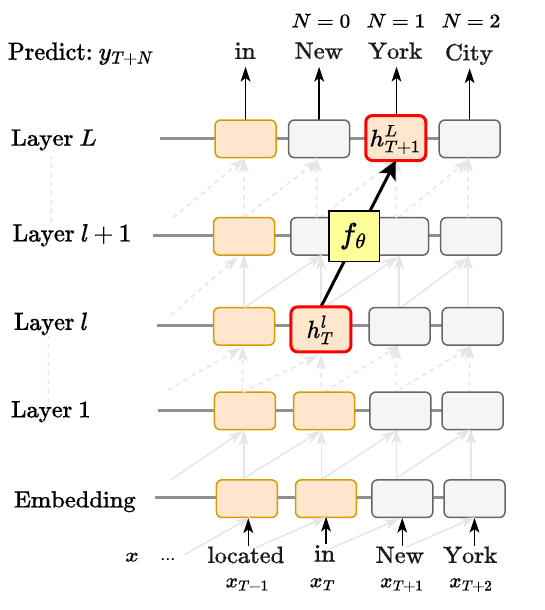}
    \caption{LLM to Linear Model Approximation Overview. Given a hidden state, $h_{T}^{l}$, the linear model, $f_\theta$, is trained to output a future hidden state $h_{T+1}^{L}$. In this example, $h_{T}^{l}$ is the encoding that would lead to the prediction of `New,' and $f_\theta$ uses only that information to predict $h_{T+1}^{L}$ that would predict `York.'}
    \label{fig:linear-model}
\end{figure}

\noindent When generating a sequence of tokens beyond the given starting prefix of length $T$, we %
write:
\begin{align}
    y_{T+i} & = G([x_1, .., x_{T+i-1}, x_{T+i}]) \\
    x_{T+i+1} & = \argmax y_{T+i}
\end{align}
Our goal is to devise methods that can anticipate what $G$ will predict for $y_{T+1}$ through $y_{T+N}$ from only a single hidden state at $h_{T}^{l}$.

\subsection{Direct Vocabulary Prediction}

Let $h_{T}^{l}$ denote the hidden representation induced by $G$ for token $x_T$ at intermediate layer $l \leq L$, 
and let $y_{T+N}$ denote the subsequent-token distribution predictions produced by $G$ after token $x_{T+N}$. To predict $y_{T+N}$ from $h_{T}^{l}$ alone, we 
train a linear model $g_\theta$ to predict logits $\hat{z}_{T+N}$ that approximate $\hat{y}_{T+N}$ after softmax:
\begin{align}
    \hat{z}_{T+N} & = g_\theta(h_{T}^{l}) \\ \nonumber
    \hat{y}_{T+N} & = \mathrm{softmax} (\hat{z}_{T+N}) \approx \hat{y}_{T+N}
\end{align}
Since this model directly predicts the subsequent predictions over the full vocabulary from $h_T^l$, we call it the direct vocabulary prediction model.

\subsection{Linear Model Approximation}

We also test a linear model based on the tuned logit lens \citep{belrose2023eliciting,din2023jump} approach, which anticipates future hidden states within the transformer and decodes them using the pretrained decoder head.  Differently from that work, we model hidden states at future tokens in rather than only at later layers.

Beginning with the hidden representation $h_{T}^{l}$, we create a model to predict a hidden state $h_{T+N}^{L}$ at the final layer $L$, and subsequent token $x_{T+N}$.  To predict $h_{T+N}^{L}$ from $h_{T}^{l}$, we %
train a linear model: %
\begin{align}
    \hat{h}_{T+N}^{L} & = f_\theta(h_{T}^{l})  \approx {h}_{T+N}^{L}
\end{align}
The vocabulary can be read from the predicted $\hat{h}_{T+N}^{L}$ by applying the pretrained decoder head of the transformer.  In Figure~\ref{fig:linear-model}, we show an example of one such linear model. Suppose that we have trained a linear model parameterized by $\theta$, $f_{\theta}$, that takes in the last token hidden representation of the input at layer $l$ to generate a hidden state at layer $L$ of the following token hidden representation. When we input the following in $G$: ``Madison Square Garden is located in", we get ``New" as the highest-probability prediction at $N=0$ and ``York" at $N=1$. 
We use the linear model to approximate this based on the hidden  representation of $T_N$ (i.e., ``in") at layer $l \leq L$ as our input;
the ideal output of the linear model given this would be the hidden state at $T_{N+1}$ and layer $L$,  which is associated with predicting ``York'' as the most probable token.

This approach differs from the direct vocabulary approach by reusing the pretrained decoder head of the transformer. We find that this marginally aids predictions at the latest layers $l$ near $L$. 
Based on the observation that other pretrained transformer parameters may encode memorized calculations that facilitate decoding of subsequent tokens, we next turn to other approaches that utilize larger portions of the pretrained transformer to predict future tokens.

\subsection{Fixed Prompt Causal Intervention}

The next method we consider involves a single-state causal intervention where we transplant the hidden state $h_T^{l}$ into the transformer while it is decoding an unrelated bit of context. 
The question is whether this transplantation steers the model to generate tokens related to the prefix that induced $h_T^{l}$. 
If it does, this indicates that information about subsequent tokens (in the original sequence) is prominently encoded in $h_T^{l}$.

\begin{figure*}
    \centering
    \includegraphics[scale=0.7]{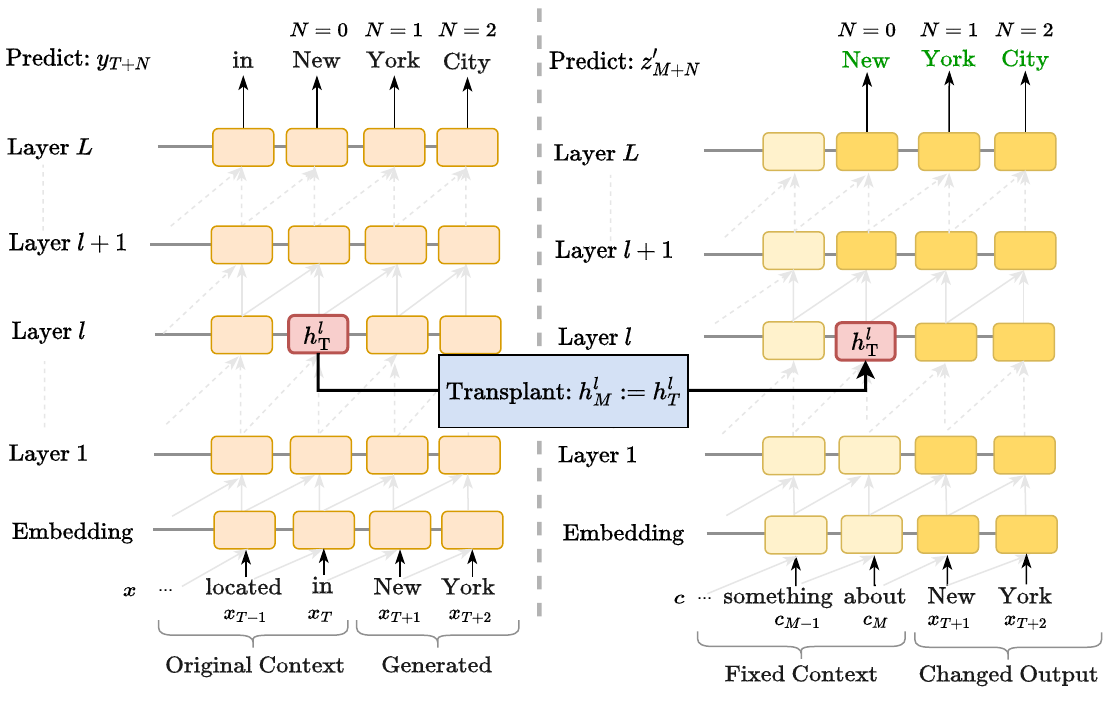}
    \caption{Illustration of Fixed prompt Causal Intervention. The left and right sides represent two different transformer model runs. On the left hand side, we have the original run of \emph{Madison Square Garden ... in New York}. We transplant the hidden state, $h_{T}^{l}$ to the other transformer model run, which has a fixed generic context, \emph{Tell me something about}, as its input. With $h_{T}^{l}$ replacing the hidden state at $h_{M}^{l}$, we measure the tendency of this modified transformer run to reveal the probability distribution in $h_{T}^{l}$. In such cases, it would reveal that $h_{T}^{l}$ was predicting, for instance, `New York City.'}
    \label{fig:causal-intervention}
\end{figure*}

Figure~\ref{fig:causal-intervention} depicts the procedure.  
On the left, we show the original context from which $h_T^{l}$ is read; here $x = [x_1, ..., x_T]$ is ``Madison Square Garden is located in" where $x_1$ is ``Madison" and $x_T$ is ``in".
This results in a sequence of outputs $[x_{T+1}, ..., x_{T+N}]$ which will read ``New York City.''  
On the right, we run a %
single generic fixed-context prompt $c = [c_1, ..., c_M]$ (e.g., ``Please, tell me something about" where $c_1$ is ``Please" and $c_M$ is ``about") through the transformer.  %
One would not anticipate that this generic prompt %
would cause the transformer to predict ``New York City''.

Using an intervention, we now directly test that hypothesis that a single hidden state at layer $l$ and token $T$ within the original run contains the information necessary to predict subsequent tokens.  We transplant the original run's state vector $h_T^l$ into the corresponding location $h_M^l$ in the fixed-context run, then allow the transformer to proceed. If the necessary contextual information is present in the new run, the resulting tokens generated would become ``New" for the current token generation and ``York" and ``City'' for the subsequent token generations.

Formally, let the sequence $x = [x_1,...,x_T]$ denote an input context that causes the model to subsequently generate $[x_{T+1}, ..., x_{T+N}]$, and let and $c = [c_1, ..., c_M]$ represent a generic fixed-context prompt where $T$ and $M$ represent the lengths of the original and fixed input prompts, respectively. 
When each are passed through $G$, we get the following predicted distributions:
\begin{align}
    y_T & = G(x)  \in [0,1]^{|V|} \\ \nonumber
    \hat{y}^*_M & = G(c)  \in [0,1]^{|V|}
\end{align}
Denote the intervention that replaces $h_M^{l}$ from the fixed-context run with state $h_T^{l}$ from the original run as:
\begin{align}
    \hat{y}_M = G(c \,||\, h_M^l := h_T^l)
    \label{eq:fixed-intervention}
\end{align}
If, after the intervention, the new predicted distribution $\hat{y}_M \approx y_M$ approximates the prediction in the original context, that will reveal that $h_{T}^{l}$ specifically encodes information needed for that prediction.

Furthermore, we can deduce what $h_{T}^{l}$ encodes about subsequent token predictions $n$ steps ahead by adding the generated tokens to the input and comparing the following predictions:
\begin{align}
\label{eq:fixed-intervention-2}
   y_{T+i} & = G(x + [x_{T+1},...,x_{T+N}])  \\ \nonumber
   \hat{y}_{M+i} & = G(c + [x_{T+1},...,x_{T+N}] \,||\, h_M^l := h_T^l ) 
\end{align}

The context prompt $c$ could be chosen as any sequence of tokens.  In practice, some prompts are more amenable to this intervention than others.  In our experiments, we will test a small set of highly generic phrases.

\begin{figure*}
    \centering
    \includegraphics[scale=0.7]{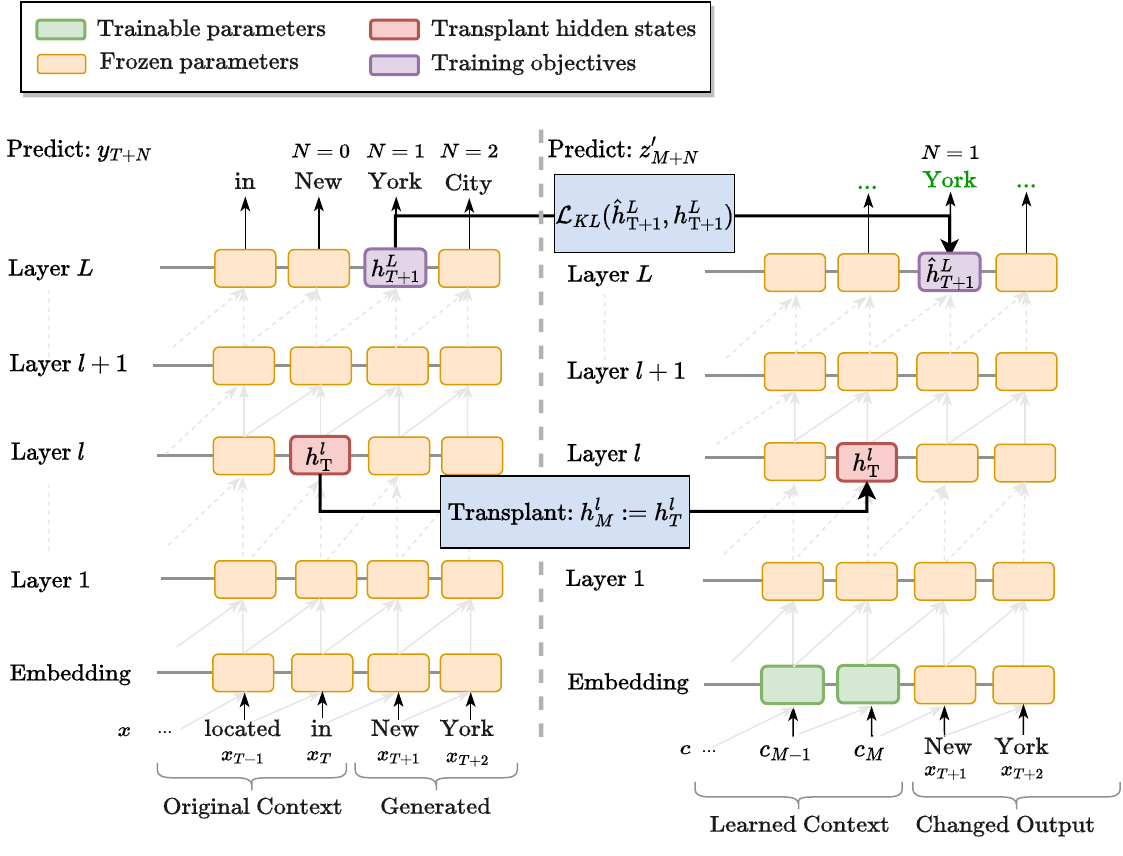}
    \caption{Learned context prompt Causal Intervention Overview. The left and right sides represent two different transformer model runs. The general setup is the same as Figure~\ref{fig:causal-intervention}. The difference lies in the context provided in the transformer run on the right hand side. Instead of manually thinking of a context, we provide a learned context to increase the tendency of decoding the subsequent tokens predicted by $h_{T}^{l}$. We do so by training the context, $c$, with $L_{KL}$ criterion and the objective to match the subsequent token prediction, such as `York' in this instance.}
    \label{fig:learned-intervention}
\end{figure*}

\subsection{Learned Prompt Causal Intervention}

In the previous section, we have described an intervention that could reveal information predictive of upcoming tokens encoded in a single hidden state, by steering generation when grafted into completely unrelated contexts. 

However, in cases where this ``fails'', it does not necessarily mean that the hidden state does not encode similar information; it may just be less prominent. To evaluate the degree to which such signal is present in these cases, we next explore an approach in which we \emph{learn} to surface information about subsequent tokens from individual contextual token embeddings. This procedure is shown in Figure~\ref{fig:learned-intervention}.

Specifically, we optimize a parameterized prefix, $c_{opt} = [c_1, ..., c_M]$ to extract this information from the hidden state. 
For each decoder layer $l$, we train the corresponding prefix $c_{\text{opt}}^{(l)} = [c_1^{(l)}, ..., c_M^{(l)}]$ to maximize the probability of the model yielding the exact subsequent phrase after the original context. 
In particular, we conduct the same causal intervention in the hidden states $h_{T}^l$. 
We then optimize the probability distribution of the subsequent generation under the learned context to be the same as the original model when all its previous generation is given correctly:
\begin{equation}
    \argmin \mathrm{KL}(\hat{y}_{M+N}\,;\, y_{T+N}) \\ 
\end{equation}
Where the predicted distribution $\hat{y}_{n}$ is given using the same intervention as described in Eq.~\ref{eq:fixed-intervention-2}:
\begin{align}
\nonumber 
    \hat{y}_{M+n} = G([c_1, ..., c_M, x_{T+1},.., x_{T+N}] \\
    \, || \, h_M^l := h_T^l)
\end{align}

We hence optimize this objective with the model frozen and only prefix left to be trained. Notably, our approach is different from the implementation of prefix tuning \cite{li2021prefix} in the sense that we back-propagate the gradient through the model instead of a temporary MLP, as empirically it produces a significantly better optimized context.

%% file: sections/03_experiments.tex
\section{Experiments and Results}

\subsection{Data}

We perform evaluation on samples of the Pile~\cite{gao2020pile}, which is the 825GB %
dataset used to train GPT-J-6B~\cite{wang2021gpt} as well as other LLMs. %

To train the linear models, we sample 100{,}000 tokens that have an average of $518$ sized-context. Amongst the 100{,}000 token samples, we use 10{,}000 of them to train for our learned prompt experiment. For testing our methods, we sample another 1000 tokens that have an average previous context length of $535$.  
To simplify our analysis of the degree to which single hidden token representations encode subsequent $n$-grams, we 
draw our samples from contexts in which the original transformer model made a correct prediction.

More specifically, %
we randomly sampled train and test data points from the subset of token locations where the autoregressive transformer under consideration correctly predicts the following token. 
In Table~\ref{tab:data_stats}, we break down the types of tokens present in the testing data by categorizing the last token ($T$) of the prefix as well as the generated tokens outputs of GPT-J , through greedy (argmax) decoding, at $N=0,1,2,3$ %
with respect to various properties, such as whether they are lower-cased tokens that start with a space, or are numerical tokens, and so on.

\begin{table*}[t]
\small
\centering
\ra{1.3}
\begin{tabular}{p{1.2in} p{0.8in} l l l l l}
\toprule
\textbf{Properties} & \textbf{Last Original Context Token} & \textbf{N = 0}  & \textbf{N = 1}  & \textbf{N = 2}  & \textbf{N = 3} & \textbf{Examples} \\
\hline
\textbf{Lowercase No Space} & 12 & 14.5 & 18.1 & 13.1 & 13.4 & `itability', `aka', `ension' \\
\textbf{Lowercase With Space} & 42 & 39.1 & 37.1 & 38.4 & 36.7 & ` sense', ` tests', ` punitive' \\
\textbf{Uppercase No Space} & 2.4 & 2.7 & 2.2 & 2.8 & 1.6 & `V', 'TABLE', 'SE' \\
\textbf{Uppercase With Space} & 1.9 & 2.4 & 1.1 & 1.5 & 1.7 & ` STAR', ` UK', ` USA'  \\
\textbf{Token length $< 4$} & 57.8 & 59.8 & 64.3 & 59.9 & 63.2 & `*', `ate', `</' \\
\textbf{Token length $\geq 4$} & 42.2 & 40.2 & 35.9 & 40.5 & 37 & ` validation', ` Subaru', `ulsion' \\
\textbf{Punctuation} & 15.7 & 14.5 & 17.3 & 15.2 & 19 & `-', `.', `</' \\
\textbf{Numerical} & 2.4 & 2.7 & 1.9 & 3.2 & 2.8 & `1998', `001', `5' \\
\hline
\end{tabular}
\caption{Data Frequency of different token properties on the Last Prefix Tokens and GPT outputs at N=0,1,2,3. \\ Each number in the table is a percentage of the test dataset, which is of size 1000.}
\label{tab:data_stats}
\end{table*}

\subsection{Evaluation Metrics} For evaluation we adopt the same metrics used in prior related work ~\citet{din2023jump}, namely Precision$@k$ and Surprisal. 

Precision$@k$ measures the appearance of the top probability token in the output at $N$ tokens ahead we predict from the hidden state 
with respect to the observed top-$k$ tokens
from GPT-J-6B model output. 
Higher values are better here because  
these mean the actual token at the corresponding future token was accurately predicted.

Surprisal, on the other hand, is the minus log probability according to the GPT-J-6B model output of the highest probability token according to the proposed probing methods. 
Lower is better for this measure because such values imply that the top predicted tokens are deemed probable by the model.

\subsection{Experimental Setup}

\paragraph{Linear Model}
We train two types of linear models --- one with %
an output space of 4096 (the hidden representation %
size used by GPT-J-6B), and the other one with 50{,}400 (the vocabulary space of the same). GPT-J-6B comprises 28 layers.
We train 4 instances for each of these layers, one for each different ``future'' token position we consider ($n=0,1,2,3$). As input we accept the source hidden state, i.e., $h_{T}^{l}$.  Our output is either the hidden state, i.e., $h_{T+N}^{L}$ or the decoded output at the position (vocabulary distribution) $T+N$.

\paragraph{Fixed Prompt Causal Intervention}
This is an evaluation-only setup where we choose four generic context prompts and perform causal intervention on these contexts as shown in Figure~\ref{fig:causal-intervention}. The four fixed context prompts that we test are:
\begin{itemize}[itemsep=0pt,parsep=1pt,topsep=3pt,leftmargin=10pt]
\item \texttt{\small Hello! Could you please tell me more about "}
\item \texttt{\small The multi-tokens present here are "}
\item \texttt{\small The concepts in this hidden state listed are: (}
\item \texttt{\small <|endoftext|> This state is describing about the following concept:}
\end{itemize}
The hidden states are gathered from layer $l$ of the last token of the context tokens and are transplanted into the hidden representation of the last token in the generic prompts at the same layer $l$.

\begin{figure*}[t]
    \centering
    \includegraphics[width=155mm]{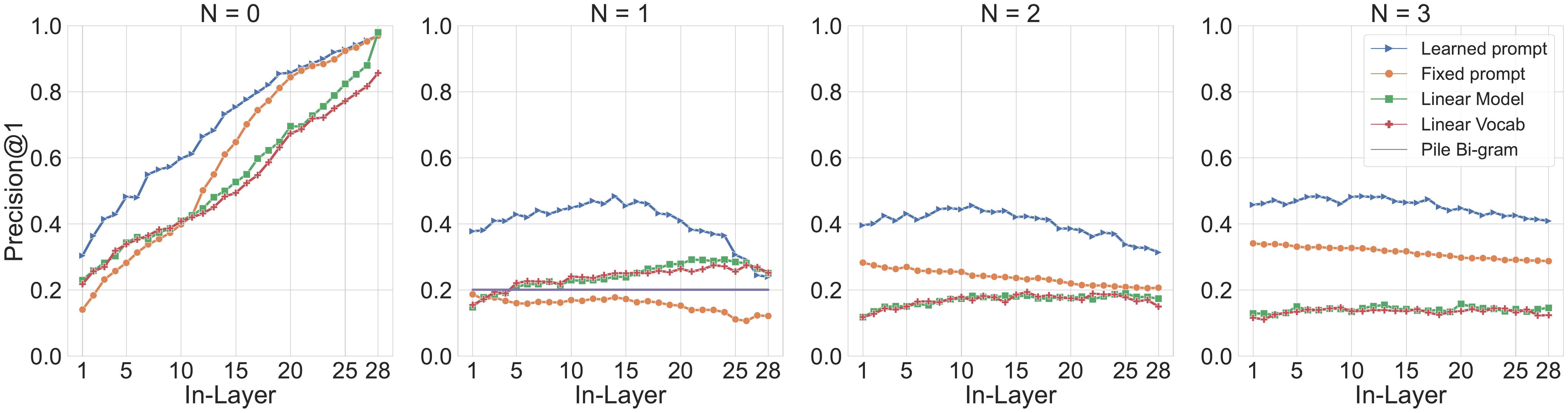}%
    \vspace{-3pt}%
    \caption{Accuracy (Precision@1) using the transplanted hidden representation. The $N=0$ case models immediate next-token prediction, and $N\geq 1$ are the subsequent-token cases that are the focus of our work. The learned prompt is best able to recover future token information from hidden states of a preceding individual token, with predictive accuracy peaking at middle layers, with more than double the accuracy of a bigram baseline. A linear model predicting the hidden state fares comparably to predicting directly into the output vocabulary.} %
    \label{fig:prec_at_k}
\end{figure*}

\begin{figure*}[t]
    \centering
    \includegraphics[width=155mm]{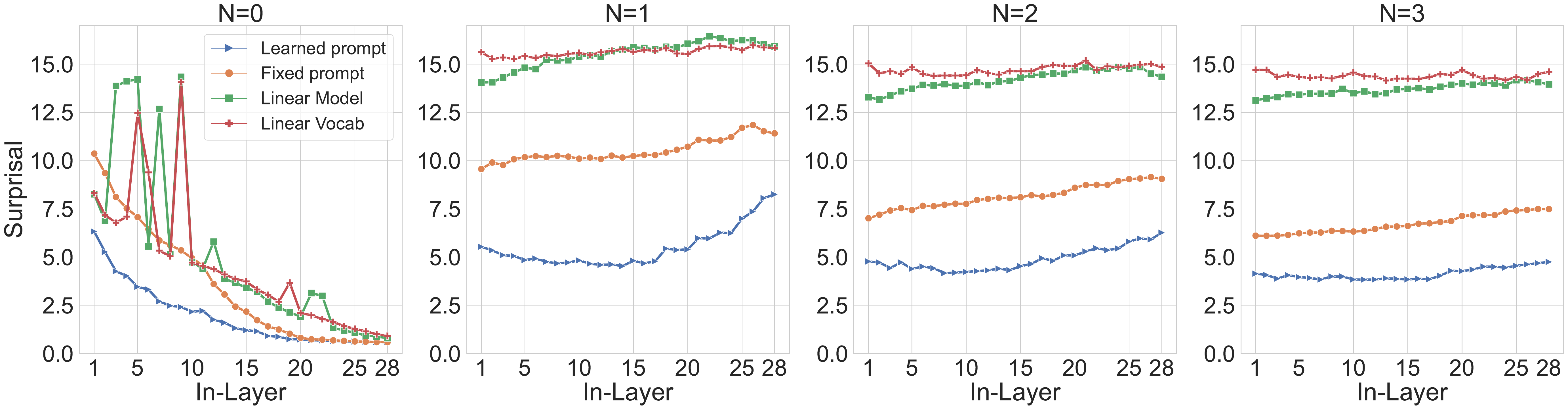}%
    \vspace{-3pt}%
    \caption{Average surprisal of the model after transplantation.  Again the learned prompt performs best, confirming the presence of subsequent-token information encoded at middle-layer hidden states.}
    \label{fig:suprisal}
\end{figure*}

\paragraph{Learned Prompt Causal Intervention}
We then compare with trained prompts with the same token length as the fixed prompts.  We train a soft prompt for each layer $l$ from 1 to 28.  Each learned prompt is trained by maximizing the probability of generating the token from the prefix context at the penultimate layer, when the hidden state is transplanted at layer $l$ at the last token of the soft prompt, in the same way as the fixed prompts are applied. %
We train a prefix with a length of 10. %
This method performs best and is our main method.

\subsection{Unveiling Subsequent Tokens}

Figure \ref{fig:prec_at_k} and Figure \ref{fig:suprisal} illustrate the difference between our method and the baselines. The learned prompt optimized with the objective of predicting the next token (N=1) has the best performance. On average, the precision@1 is 24.8\% higher, precision@5 is 25.3\% higher, and precision@10 is 25.1\% higher than the \textbf{best} baseline method. The bigram baseline\footnote{The bigram baseline is collected from 900{,}000 documents from the Pile dataset.} at N=1 is shown as a horizontal line; the bigram model i achieves 20.1\% accuracy.
For surprisal, the learned prompt also has the lowest value, which indicates its efficacy at maximally unveiling the information behind the hidden states.

\begin{table}[t]
    \centering
    \begin{tabular}{l c c c c}
        \toprule
         & \textsc{Lens} & N=1 & N=2 & N=3 \\
         \midrule
         \textbf{Accuracy} & \\
        \textsc{Learned} & 97.0 & \textbf{48.4} & \textbf{43.7} & \textbf{46.9} \\
        \textsc{Fixed} & 97.0 & 20.8 & 30.0 & 36.5 \\
        \textsc{HS} & \textbf{98.0} & 29.2 & 19.0 & 15.8 \\
        \textsc{Vocab} & 85.7 & 27.5 & 19.4 & 14.7 \\
        \midrule
        \textbf{Surprisal} & \\
        \textsc{Learned} & \textbf{0.6} & \textbf{4.5} & \textbf{4.4} & \textbf{3.9} \\
        \textsc{Fixed} & \textbf{0.6} & 8.8 & 6.5 & 5.7 \\
        \textsc{HS} & 0.8 & 14.1 & 13.2 & 13.1 \\
        \textsc{Vocab} & 0.9 & 15.3 & 14.4 & 14.2 \\
        \bottomrule
    \end{tabular}
    \caption{Best accuracy and surprisal results for each method. \textsc{Learned} refers to the Learned Prompt Causal Intervention Method; \textsc{Fixed} denotes the Fixed version. \textsc{HS} is the Linear Model variation that predicts Hidden State;  \textsc{Vocab}, %
    is the Linear Model variation that predicts a distribution over the vocabulary directly.}%
    \label{tab:best-acc}
\end{table}

\begin{table*}[hbt!]
\small
\centering
\ra{1.3}
\begin{tabular}{p{1.3in} p{1.3in} l l l}
\toprule
\textbf{Last Context Token Type} & \textbf{Linear: Vocab Space} & \textbf{Linear: Hidden State}  & \textbf{Fixed Context}  & \textbf{Learned Context} \\
\hline
\textbf{Lowercase No Space} & 21.7 & 25.2 & 9.2 & \textbf{32.5} \\
\textbf{Lowercase With Space} & 26.4 & 20.8 & 19.2 & \textbf{51.9} \\
\textbf{Uppercase No Space} & \textbf{29.2} & 26.3 & 0.0 & \textbf{23.3} \\
\textbf{Uppercase With Space} & 26.3 & 26.3 & 10.5 & \textbf{31.6} \\
\textbf{Token length $< 4$} & 26.5 & 24.9 & 21.8 & \textbf{46.9} \\
\textbf{Token length $\geq 4$} & 23.9 & 24.4 & 18.0 & \textbf{52.1} \\
\textbf{Punctuation} & 28.7 & 28.7 & 16.6 & \textbf{47.8} \\
\textbf{Numerical} & 12.5 & 16.7 & 20.8 & \textbf{33.3} \\
\hline
\end{tabular}%
\vspace{-1pt}%
\caption{Accuracy of predicting $N=1$ token ahead ($y_{T+1}$, which predicts $x_{T+2}$) based on hidden representation of the last context token($x_T$).  Results are shown for layer $l=14$, where the learned prompt model is most accurate.}
\label{tab:n-1-accuracy}
\vspace{-2pt}%
\end{table*}

\subsection{Contexts of Accurate Predictions}
To further explore the contexts in which these methods seem better (or worse) able to predict subsequent tokens, we categorize input token (the last original context token) into eight (non-mutually exclusive) categories, shown in Table~\ref{tab:n-1-accuracy}.  
We report the model accuracies when using layer 14, where the learned prompt model peaks.

While all categories of token types are predicted better by the learned prompt than by the linear model, the relative improvement is highest when the last context token is a lowercase token preceded by a space, or a longer token. This suggests that information about how to complete long words may not be immediately accessible by a linear model decoder, but that they can be made accessible by using the parameters of the pretrained model as done by the learned prompt intervention method. 

We have also observed that the accuracy of predicting subsequent tokens is 
correlates with the model's confidence in its next token prediction.
In the case of $N=1$, for instance, the learned prompt intervention method's calibrated accuracy is 26\%, 57\%, 77\%, and 95\% for model confidence groups of 0-30\%, 30-60\%, 60-90\%, and 90\%-100\%, respectively. 
These trends appear in $N = 2$ and $N = 3$ as well. 
This suggests that we might gainfully use this decoding method as a probing tool, trusting that predicted future tokens are generally accurate when the model is confident.

Does future information appear only in the presence of higher-level concepts?  For example, one might hypothesize that in cases the language model predicts an entire named entity, that the probing method might decode future predictions more accurately.  To investigate this, we performed sub-group analyses on test results to characterize how well the best probing method performed specifically for multi-token named entities. Interestingly, we found little difference: when examining just the named entity cases, we observe similar or slightly lower accuracy: 44\%, 42\% and 37\% for $N = 1,2,3$, suggesting that future information is present broadly, not only for long entity names.

In sum, we have found that a single hidden state encodes information about outputs more than one token ahead, and we have demonstrated three different methods that can decode them for GPT-J-6B. 

\input{sections/future-lens-figure}
\paragraph{Application: Future Lens}
We apply Learned Prompt Intervention to create a novel probing tool we call the \textit{Future Lens}. 
Given a soft prompt, we perform the intervention using the states arising from the user's prompt to provide a view into what the hidden states encode about future tokens. %
In Figure~\ref{fig:future_lens_example}, we show  
an example %
for the prompt: 
``Marty McFly from".
The Future lens %
reports the anticipated four tokens from every hidden state in the model (across layers). %

In the Future Lens visualization, every cell represents a hidden state from a particular layer ("L\{digit\}") at a specific token. The shade of each cell indicates the model's average confidence with respect to the corresponding token predictions (darker shades indicate greater confidence).  For example, at the cell representing the hidden state at Layer 25 at the token ``from", we can see that the confidence in the predicted tokens ``Back to the Future" is strong. This particular state suggests that the LLM already knows that Marty McFly is related to the Back to the Future movie. 
Interestingly, the model also assumes ``Marty'' to have the surname Donough. 
Returning to the predictions at token ``from", we see that the early layers seem to first predict locations such as Australia or Boston. However, through future predictions, we can see the model begins to associate Marty McFly with a movie around Layer 6. Hence, through this tool, we can gain further insights about the model's chain of predictions at every hidden state. 

%% file: sections/future-lens-figure.tex
\begin{figure*}[h]
    \centering
    \includegraphics[width=0.94\linewidth]{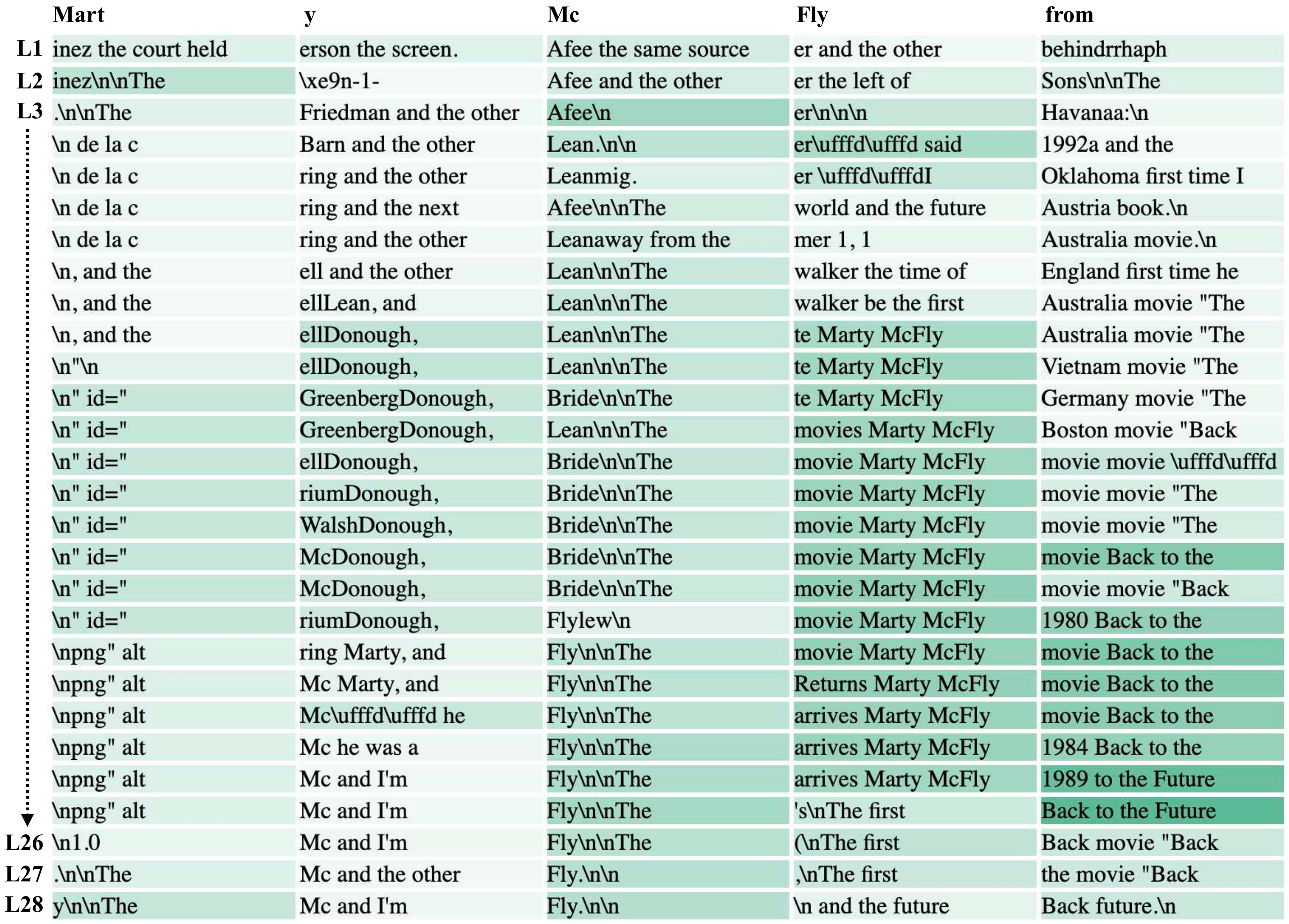}%
    \vspace{-1pt}%
    \caption{The Future Lens applied to the hidden states of GPT-J-6B processing \emph{Marty McFly from}. Each cell illustrates the most likely sequence of future tokens that the respective hidden state predicts. The darker boxes correspond to higher probabilities/confidence.}
    \vspace{-2pt}%
    \label{fig:future_lens_example}
\end{figure*}

%% file: sections/05_relatedworks.tex
\section{Related Work}

\paragraph{Knowledge Prediction and Manipulation} Recent works have %
delved into %
LLM internals %
to better understand how %
such models predict the next token at each computation step. Geva \emph{et al.} \citeyearpar{geva-etal-2021-transformer}, for instance, find that %
the feed-forward layers in transformers operate as key-value memories, %
allowing one to intervene %
at those layers to modify the next token output~\cite{geva-etal-2022-lm}. Frameworks such as ROME~\cite{meng2022locating} and MEMIT~\cite{meng2022memit} scale such manipulations to edit knowledge in stored in LLMs. 

The consensus %
that has emerged in these papers %
is that %
some early-middle and late layer calculations contribute the most to the final predicted token. 
Tools such as %
Logit lens~\cite{logit_lens} and Tuned lens~\cite{belrose2023eliciting,din2023jump} %
allow us to look at the top-$k$ values of the transformer at \emph{every} layer and token to see early next-token predictions.
Katz and Belinkov ~\citeyearpar{katz2023interpreting} used logit lens to visualize semantic information flow in GPT-2 models. While these works primarily deal with next-token predictions, ~\citet{hernandez2023linearity} shows that specific attributes of an entity can be extracted with an affine transformation on the entity representation long before the LM is actually required to predict the attribute, enabling an ``attribute lens'' on early layers and early tokens. We aim to characterize how the current hidden state would affect the prediction of both the next token and tokens farther ahead, but unlike ~\citet{hernandez2023linearity}, we deal with open contexts and are not constrained to certain relations. 

\paragraph{Early Exit Decoding} To optimize the running time and space requirements of training models, %
prior work has looked at ``early exit'' strategies, which usually involves stopping at earlier layers of computation and estimating the final predictions based on those computations~\cite{schuster2022confident, xin-etal-2021-berxit, kong-etal-2022-accelerating, zhang2020accelerating, din2023jump}. The takeaway from these methods is that it is possible to achieve %
prediction performance comparable to %
that observed when all layers are used even when dropping a couple of computational layers %
for each token. 
For instance, Din and colleagues \citeyearpar{din2023jump} %
used linear transformations to predict a later layer's hidden representation from an earlier layer at the same token. 
This approach was able to preserve $\sim$95\% of the %
full transformer model outputs on GPT-2~\cite{radford2019language} and BERT~\cite{bert}. 
This result implies that initial model layers encode information that to largely determines the final output.
In this work we test 
the limits of this phenomenon by evaluating the degree to which a single hidden state for a token at position $T$ can be used to predict tokens multiple steps ahead (i.e., at $T+N$).

\paragraph{Memorization in Language Models} Due to the %
potentially sensitive %
information present in %
the datasets used to train language models (LMs), past work %
has investigated %
what, when, and why memorization occurs~\cite{carlini21extracting, thesecretsharer19, feldman2020,lehman2021does}, how memorization changes as %
a function of training data size ~\cite{carlini2023quantifying, wei2022emergent}, and how other memorized information can be detected based on model internal states ~\cite{haviv-etal-2023-understanding}. 

These works have %
collectively illustrated that there are some text snippets that LMs %
remember and can output %
verbatim or in closely paraphrased versions (``approximate memorization''; ~\citealt{ippolito2023preventing}). 
Other work ~\cite{haviv-etal-2023-understanding} has shown that earlier layers of models tend to promote memorized concepts or tokens, while %
later layers %
boost %
model confidence %
in these tokens. 
Our paper can be viewed as an extension of this work on investigating memorization of multi-token phrases: we ask whether and to what extent a single model hidden state encodes multi-token information.

\paragraph{Prompt Tuning}

Prompt Tuning %
has emerged as a parameter-efficient method for %
fitting LMs %
for new downstream tasks. By freezing the %
LM and optimizing only the soft prompt parameters, models are able to achieve performance comparable%
to that observed after fine-tuning all parameters. %
Li \emph{et al.} \citeyearpar{li2021prefix} introduced %
prefix tuning which entailed training plug-and-play prefix that steers the behavior of the LMs for the downstream tasks. 
Other work \cite{wallace2019universal} applied a gradient-based method to search for the best discrete prompts which enable the model to produce desire generation. 
Sun and colleagues \citeyearpar{sun2023evaluating} train the prefix soft prompt as a way of aligning %
semantically equivalent instructions in latent space.

%% file: sections/04_discussion.tex
\section{Discussion}

In this paper, we explored the degree to which we are able to decode multi-token outputs subsequent to a particular token on the basis of its hidden representation alone.  
The results in Table~\ref{tab:best-acc} and Figures~\ref{fig:prec_at_k} and~\ref{fig:suprisal}  indicate that the representations encode such information, to some degree.
Among the decoding methods we assessed, learned prompts are best able to predict such future tokens.  
Both the linear and the learned prompt models achieve better accuracy than the empirical bigram baseline at $N=1$.
Interestingly, predictive accuracy of the learned prompt model peaks at the middle-layer hidden states, suggesting that subsequent-token information is encoded at those layers; this pattern is very different from the immediate next-token $N=0$, in which accuracy peaks at the last layer.

The learned prompt model realizes an accuracy sufficient to be potentially useful as a ``Future Lens'' to provide insights about subsequent token information contained in hidden states within LLMs.
This provides a way to decode a short sequence of tokens encoded in a hidden state, rather than only the single immediate token prediction.

{\paragraph{Data and Code Availability}
All code and data for demo and implementation is made available at: \url{https://future.baulab.info}}

{\paragraph{Acknowledgements}
This work was supported by Open Philanthropy and by the National Science Foundation (NSF) award 1901117.  We thank the Center for AI Safety (CAIS) for making computing capacity available for this research.}

%% file: sections/Appendix.tex
\onecolumn
\section{Appendix}
\label{section:appendix}

\section*{Additional Figures}

In this main paper, we report results based on models that are trained to optimize the $N=1$ single token-ahead prediction, and we test those models for predictive accuracy for other $N$.

The same methods can also be used to optimize subsequent tokens, and the results of those methods are shown here.  We find that optimizing for $N=1$ works best and generalizes surprisingly well to other $N$, but that that optimizing for other $N$ does not perform well for $N=1$.

\begin{figure*}[h]
    \centering
    \includegraphics[width=155mm]{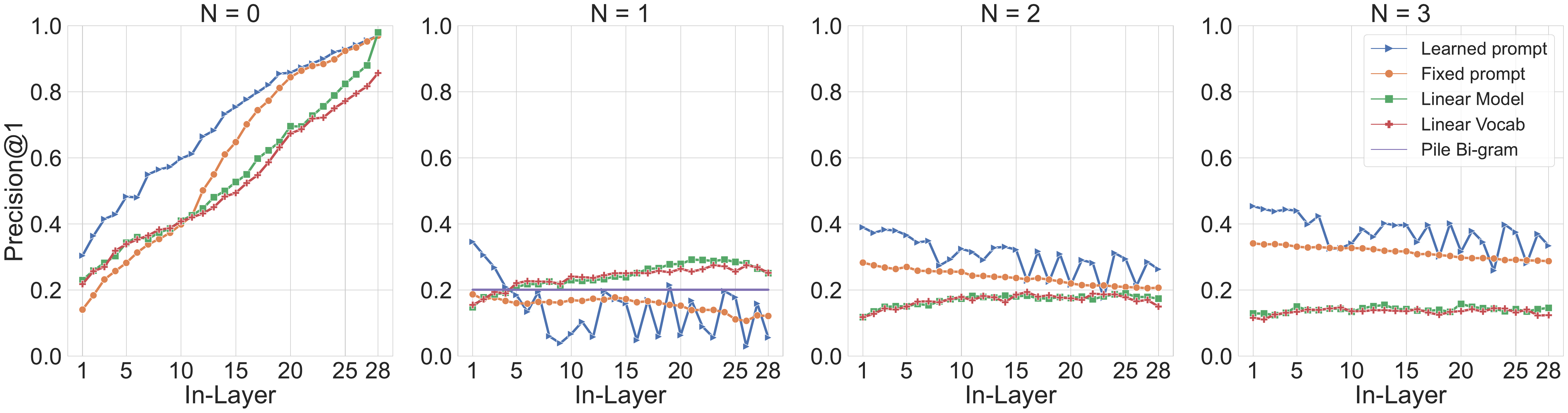}
    \caption{The Precision@1 (Accuracy) of all the methods trained with predicting the currently decoded token (teacher-forcing)}
    \label{fig:appendix_precision_n=0}
\end{figure*}

\begin{figure*}[h]
    \centering
    \includegraphics[width=155mm]{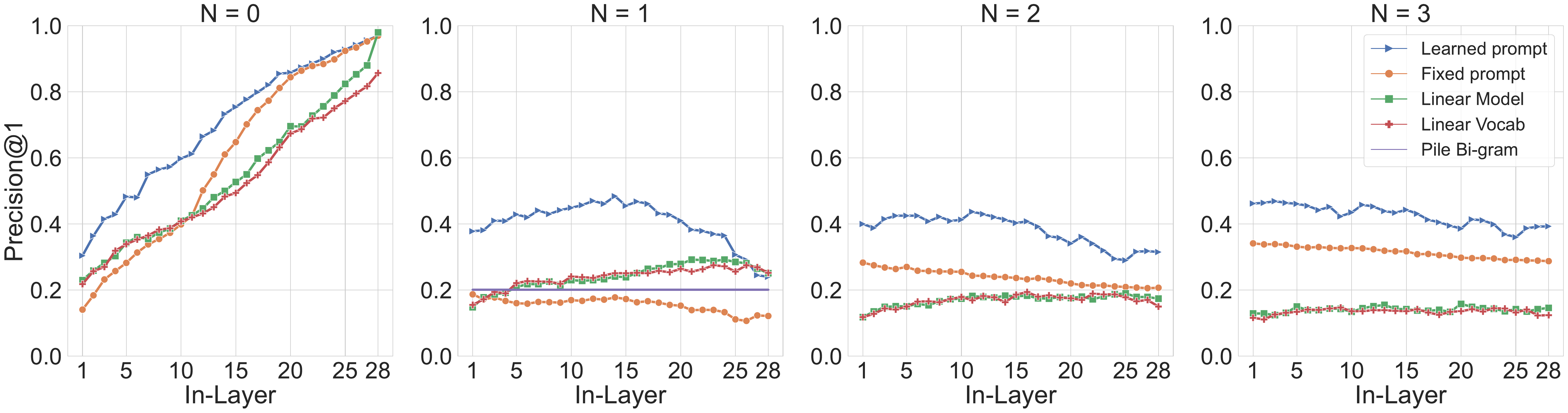}
    \caption{The Precision@1 (Accuracy) of all the methods trained with predicting the 1st next token}
    \label{fig:appendix_precision_n=1}
\end{figure*}

\begin{figure*}[h]
    \centering
    \includegraphics[width=155mm]{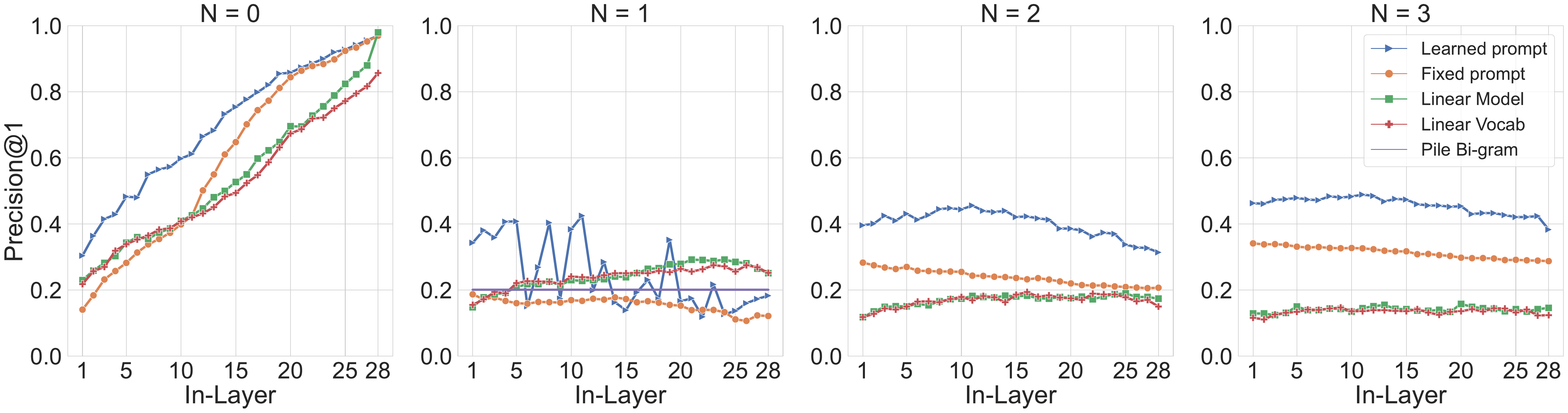}
    \caption{The Precision@1 (Accuracy) of all the methods trained with predicting the 2nd next token}
    \label{fig:appendix_precision_n=2}
\end{figure*}

\begin{figure*}[h!]
    \centering
    \includegraphics[width=155mm]{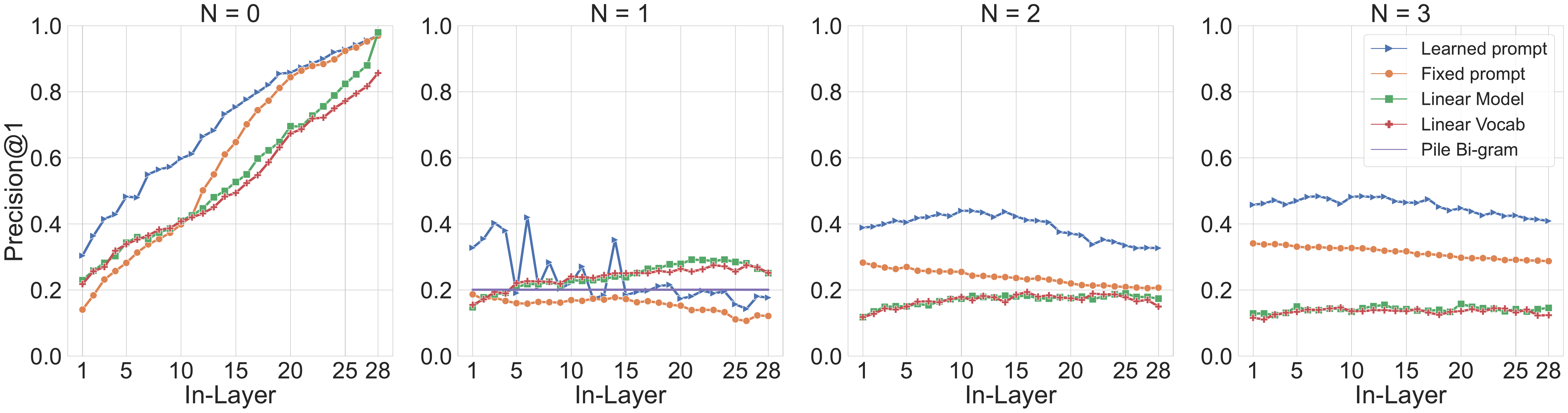}
    \caption{The Precision@1 (Accuracy) of all the methods trained with predicting the 3rd next token}
    \label{fig:appendix_precision_n=3}
\end{figure*}

\newpage
\section*{Limitations}
In our exploration with extracting far future tokens from single hidden states, we have mostly trained and tested on English data whose size, 100,000, is still relatively small compared to the data size that GPT-J-6B was actually trained in. Furthermore, the experiments were only conducted in GPT-J-6B. While the presence of subsequent token information in a single hidden state is evident in this model, it would be more comprehensive to run these experiments in other LLMs. Since there are no specific prior works that focused on decoding far future tokens from a single hidden state, we did not have any prior baselines we would refer to. While we did create a bigram baseline in the case of predicting 2 tokens in the future ($N=1$) and also create linear models as a first decoding method, there could be baselines with other architectures like Recurrent Neural Networks~\cite{jordan1997serial, ELMAN1990179} and Non-Autoregressive generation~\cite{su-etal-2021-non, xiao2023survey}. Lastly, our experiments were up to 4 tokens in the future, i.e., $N=0,1,2,3$. It would be intriguing to scale and test up to how many tokens in the future does a single state actually encode and predict.